\documentclass[preprint, 3p, twocolumn]{elsarticle}

\usepackage{lineno,hyperref}
\usepackage{amsmath,amssymb,amsfonts}
\usepackage{algorithmic}
\usepackage{graphicx}
\usepackage{textcomp}
\usepackage{soul}

\usepackage{booktabs}
\usepackage[table]{xcolor}
\usepackage{tabularx}
\newcolumntype{Y}{>{\centering\arraybackslash}X} 
\usepackage{tikz}
\usepackage{import}
\usepackage{siunitx}
\usepackage{csvsimple}
\usepackage{subcaption}
\usepackage{csquotes}
\usepackage{gensymb}
\usepackage{nomencl}
\makenomenclature
\usepackage{calc}  
\usepackage{enumitem}  

\usepackage[acronym, nonumberlist]{glossaries}
\makeglossaries

\usepackage{chemmacros}

\usepackage[version=4]{mhchem}
\usepackage{layouts}
\usepackage{pgf-pie}
\usetikzlibrary{chains}
\usepackage{multirow}

\usepackage[labelfont=bf,justification=raggedright,singlelinecheck=false]{caption}
\definecolor{sampleBlue}{rgb}{0.12, 0.47, 0.71}	
\definecolor{darkPurple}{rgb}{0.27, 0, 0.33}		
\definecolor{darkGreen}{rgb}{0.13, 0.56, 0.55}		
\definecolor{darkYellow}{rgb}{0.99, 0.90, 0.15}	
\definecolor{lightPurple}{rgb}{0.92, 0.90, 0.93}	
\definecolor{lightGreen}{rgb}{0.91, 0.95, 0.95}	
\definecolor{lightYellow}{rgb}{1, 0.99, 0.91}		

\newcommand{\new}[1]{\new{ #1}}

\newcommand\topic[1]{\textcolor{orange}{}}

\nomenclature{\(\alpha_{PV}\)}{Inclination angle of PV system}
\nomenclature{\(\beta_{PV}\)}{Orientation angle of PV system}   
\nomenclature{\(P_{PV}\)}{Peak power of PV system}        
\nomenclature{\(C_b\)}{Nominal Battery Capacity}        
\nomenclature{\(b_\mathrm{max}\)}{Maximum Battery state of charge}  
\nomenclature{\(b_\mathrm{min}\)}{Minimum Battery state of charge} 
\nomenclature{\(P_{c}\)}{Battery Charging Threshold}       
\nomenclature{\(P_{d}\)}{Battery Discharging Threshold}  
\nomenclature{\(V\)}{Heat storage cylinder volume} 
\nomenclature{\(C_\text{invest}\)}{Investment cost}
\nomenclature{\(C_\text{annual}\)}{Annual operation cost}
\nomenclature{\(R\)}{Resilience}
\nomenclature{\(G\)}{Yearly CO\textsubscript{2} emissions}
\nomenclature{\(\bar{b}\)}{Mean battery state of charge}         
\nomenclature{\(E_{d}\)}{Yearly discharged energy (from the battery)}
\nomenclature{\(P_{p}\)}{Maximum power peak}
\nomenclature{\(t\)}{Medium SOC time share} 
\nomenclature{\(E_{f}\)}{Yearly feed-in energy}
\nomenclature{\(P_{f}\)}{Maximum feed-in power peak}
\nomenclature{\(Q\)}{Concept quality metric}

\newacronym{ahp}{AHP}{Analytical hierarchy process}
\newacronym{chp}{CHP}{Combined heat and power plant }
\newacronym{mcdm}{MCDM}{Multi-criteria decision making}
\newacronym{pv}{PV}{Photovoltaic}
\newacronym{soc}{SOC}{State of charge}

\modulolinenumbers[5]

\journal{Energy Conversion and Management: X}

\begin{document}

\begin{frontmatter}

\title{Identification of Energy Management Configuration Concepts from a Set of Pareto-optimal Solutions}


\author[1]{Felix~Lanfermann\corref{cor1}}
\ead{felix.lanfermann@honda-ri.de}

\author[2]{Qiqi~Liu\corref{cor1}}
\ead{qiqi6770304@gmail.com}

\author[2]{Yaochu~Jin}
\ead{jinyaochu@westlake.edu.cn}

\author[1]{Sebastian~Schmitt}
\ead{sebastian.schmitt@honda-ri.de}

\cortext[cor1]{Corresponding author: Felix Lanfermann and Qiqi Liu}

\affiliation[1]{
organization={Honda Research Institute Europe GmbH},
addressline={Carl-Legien-Strasse 30},
postcode={D-63073},
city={Offenbach/Main},
country={Germany}}

\affiliation[2]{
organization={School of Engineering, Westlake University},
postcode={310030},
city={Hangzhou},
country={China}
}

\begin{abstract}
Implementing resource efficient energy management systems in facilities and buildings becomes increasingly important in the transformation to a sustainable society. 
However, selecting a suitable configuration based on multiple, typically conflicting objectives, such as cost, robustness with respect to uncertainty of grid operation, or renewable energy utilization,  is a difficult multi-criteria decision making problem.
The recently developed concept identification technique can facilitate a decision maker by sorting configuration options into semantically meaningful groups (concepts).

In this process, the partitioning of the objectives and design parameters into different sets (called description spaces) is a very important step.  
In this study we focus on utilizing the concept identification technique for finding relevant and viable energy management configurations from a very large data set of Pareto-optimal solutions.  
The data set consists  of \SI{20000}{} realistic Pareto-optimal building energy management configurations generated by a many-objective evolutionary optimization of a high quality Digital Twin energy management simulator.
We analyze how the choice of description spaces, i.e., the partitioning of the objectives and parameters, impacts the type of information that can be extracted.
We show that the decision maker can introduce constraints and biases into that process to meet expectations and preferences. 
The iterative approach presented in this work allows for the generation of valuable insights into trade-offs between specific objectives, and constitutes a powerful and flexible tool to support the decision making process when designing large and complex energy management systems.

\end{abstract}

\begin{keyword}
Energy Management\sep Configuration Concepts\sep Concept Identification\sep Clustering\sep Multi-criteria Decision Making
\end{keyword}
\end{frontmatter}



\section{Introduction}
Using fossil resources in an efficient way and reducing the consumption of energy to combat global warming have become ever more important.
This mandates an effective management of energy consumers and producers in building facilities, especially for larger industrial facilities, and has attracted attention towards an elaborate investigation of possible configuration options. 
However, energy management includes not only site selection~\cite{SHAO2020377}, but also modelling and optimizing building configurations~\cite{iijima2022automated, rodemann2018many, faccio2018state}, and developing strategies for optimal operation~\cite{gruber2015advanced, bui2019internal, nagpal2021optimal}. 
Among them, building management configurations play a vital part in helping to reduce energy consumption. 
Thus, it is of great importance to delve into the selection of a set of reasonable configurations for achieving effective energy management.

A multitude of factors has to be simultaneously considered regarding various configuration options for buildings and industrial facilities. 
These configuration options include, for example, investing in renewable energy production systems like photovoltaic (\acrshort{pv}) systems to reduce energy costs and greenhouse gas emissions.
Also, the option may include a well-managed battery energy storage system, since it can be used to counteract high power demand and mitigate additional stress on the electricity grid and corresponding fees from the energy supplier.
Further, a combined heat and power plant serving the thermal and electric needs and increasing independence from the electricity grid is also one of the key configuration options, which is particularly important in situations with unstable infrastructure.
Therefore, the options for building energy configurations can be various and plenty of combinations of appliances and infrastructure systems have to be considered.
Given the various configuration options, it is also challenging to evaluate the quality of different configuration combinations, as it is determined by a multitude of objectives, such as investment and regularly recurring cost, emissions, lifetime and profitability of equipment, and resilience towards unexpected contingencies.

Over the past decades, a large body of work has studied the optimization of building energy management using genetic algorithms or swarm intelligence algorithms, such as \cite{delgarm2016multi,harkouss2018multi,he2019investment}, to get the best energy management configurations; however, few of them have studied the optimization of building energy management when there are many conflicting objectives, let alone the analysis of many-objective solution sets.
Very recently, in our previous work \cite{Liu2022}, we conduct a many-objective optimization of a building energy management problem using several state-of-the-art surrogate-assisted evolutionary algorithms, though the obtained solution set is not fully analyzed.

After optimizing, neither selecting a representative solution nor extracting useful information from a large many-objective data set is an easy task, especially when the decision-makers have no clear preferences over different objectives. 
For instance, Ciardiello et al. \cite{ciardiello2020multi} select the best cases according to each criterion/objective and utopia point for analysis.
Earlier studies also use multi-criteria decision-making (\acrshort{mcdm})~\cite{triantaphyllou2000multi} to post-processing the solutions found by optimization algorithms. 
Among them, a ranking decision-making technique ELECTRE III is the most widely used and is often combined with the analytical hierarchy process.
For representative work, it can be referred to \cite{harkouss2018multi}. 

The aforementioned approaches, such as \cite{ciardiello2020multi, bejarano2023_clustering}, albeit effective, still have difficulties in facilitating decision-making. 
When typical clustering techniques are employed to identify groups of similar samples in a data set described by many conflicting objectives, the found clusters are often hard to interpret.
In addition, for problems with more than three objectives, i.e., many-objective problems, the visualization of the Pareto optimal solutions in existing work often relies on parallel coordinate plots~\cite{9209056}, which may not reflect the distribution of the solutions. 
Therefore, it is believed that extracting representative solutions from a many-objective solution set and visualizing solution distributions to facilitate decision-making are very challenging. 

This is particular true in the energy domain.
Typically, numerous conflicting objectives exist that are hard to grasp for human users.
Balancing economical, ecological, safety, and other needs as well as visualizing and communicating decisions to users are particular important and challenging.
Due to the heterogeneity of the conflicting objectives in the energy domain, we believe that it is particularly sensible to look at the data from multiple perspectives. 

In recent times, concept identification \cite{Lanfermann2020, Lanfermann2022a} has emerged as another way to analyze data sets. 
Motivated by the fact that the concept identification process not only provides insight into the problem, but also maps the available trade-offs in the configuration for the decision-makers, in this study, we propose using concept identification for post-processing a set of Pareto-optimal solutions, each solution being composed of nine decision variables and ten objectives.
The notion of description spaces allows us to split the features of the optimization data set into sub-projections that are meaningful to the user. 
There are several options of how to separate the features into different description spaces, which have not been explored and evaluated in previous works.
Therefore, a thorough analysis is subject of this work.

In the following, we present two main contributions of this work.

\begin{itemize}
    \item This work answers the question, how meaningful configuration concepts can be identified from a many-objective data set of several thousand building energy management options. 
    With the identified concepts, engineers can make more informed investment decisions. 
    \item This work clarifies how the choice of description spaces in concept identification impacts the identifiable trade-off possibilities in a general way and for the specific use-case, which could be used as a guideline for analyses in other real-world scenarios.
\end{itemize}

The remainder of this paper is structured as follows:
Section~\ref{sec:related work} lists related work from the fields of decision-making for many objectives, clustering, and concept identification.
Section~\ref{sec:choice_of_description_spaces} provides a description of the applied concept identification method, as well as a discussion on the importance of the choice of description spaces.
To answer the question, how meaningful configuration concepts can be identified from a large data set containing several thousand solutions for the described energy management configuration problem, Section~\ref{sec:results} illustrates the results of the set of experiments.
The investigated data set contains thousands of feasible energy management configuration options based on a \acrshort{pv} system, a combined heat and power plant, as well as a stationary battery.
A first experiment demonstrates the impact that the choice of description spaces has on the potentially identifiable concepts.
In a second experiment, a sensible combination of description spaces based on the given features is chosen to identify concepts of technically feasible and economically reasonable configurations.
A third experiment finalizes the selection process by identifying concepts within the remaining set of solutions from a previous concept and analyzing and evaluating the groups.
Section~\ref{sec:discussion} discusses the findings and improvement potential before Section~\ref{sec:conclusion} concludes the work and offers an outlook on future activities.

\section{Related work on decision making and concept identification}\label{sec:related work}

\subsection{Related work on decision making}
It is difficult to select a suitable solution from a many-objective data set for decision-making.
This is because the number of non-dominated solutions for many-objective problems is usually very large and it is challenging for decision-makers to make decisions based on the obtained candidate solutions.
To arise the attention of decision making in multi-objective optimization in the domain of energy systems, Jing et al. \cite{jing2019comparative} compare four posterior decision-making approaches, i.e., Shannon entropy, Eulerian distance, fuzzy membership function and evidential reasoning, to show that different decision-making approaches will result in different decisions on building energy system management.
In \cite{jing2019comparative}, it is concluded that given suitable preferences of different stakeholders, a reliable decision can be made for a multi-objective optimization of problems in the energy domain.

In real-world scenarios, decision-makers may likely have no clear preferences over conflicting objectives, making selecting the `best of best' solution among all Pareto-optimal solutions more difficult.
A usual approach when there are no preferences over different objectives is to analyze the extreme points or knee points. 
For example, Liu et al. analyze not only the best solution for each objective (called extreme points) but also two knee solutions to gain more insights into the many-objective Pareto-optimal solutions for building energy management~\cite{Liu2022}. 
Schmitt et al. further propose to select continuous knee regions based on the objective values~\cite{schmitt2022incorporating}. 
Delgarm et al. treat each objective of the same importance and a weighted sum is used to convert the three-objective objective values into a single-objective criterion for evaluating the quality of solutions, from which the solution with the best criterion value is selected~\cite{delgarm2016multi}. 
Sedighizadeh et al. \cite{sedighizadeh2018stochastic} determine the best compromised solution on the Pareto front set using a fuzzy decision maker, and the results show that this fuzzy approach outperforms the weighted sum and the min-max techniques with much lower cost and only slightly higher percentage of emissions. 
However, just observing one representative solution such as knee point or extreme point may not fully extract insightful information from a large set of Pareto-optimal solutions and may not help decision-makers have a deep understanding of the problem itself. 

Recently, data-mining methods have been applied to the energy management domain, attempting to extract the relationship between design parameters and optimization objectives for better control. 
For representative studies, it is referred to \cite{yu2022extracting, Lanfermann2022b}. 
Yu et. al. \cite{yu2022extracting} made a first attempt to study rule extraction from a set of Pareto-optimal solutions for a building control problem, by combining the strategy of DBSCAN clustering and low-dimensional classification trees to identify dominate patterns.
Apart from the aforementioned data-mining methods, it is also found that it is beneficial to group similar designs or configurations into concepts~\cite{Rosch1975}, to analyze a set of many-objective Pareto-optimal solutions~\cite{Lanfermann2022b}. 
A concept can incorporate different candidate solutions that share similar characteristics, typically in terms of their specification (design parameters), but also in terms of other features, such as operation mode and performance criteria (objective values).  
Thus, identifying meaningful concepts not only allows us to gain technical insight into the engineering task, but also maps the available trade-offs in the configuration for the decision maker.
Furthermore, the identified concepts can also enable decision-makers to select representative instances in the form of archetypal solutions, which can be further utilized in, e.g., additional refinement steps or subsequent optimization studies under changed boundary conditions~\cite{leImp2013}.

\subsection{Related work on concept identification}
Concept identification, as an unsupervised method, can provide sets of samples that are similar with respect to multiple sets of features~\cite{Lanfermann2020}.
It differs from clustering algorithms, such as subspace clustering~\cite{Sim2013}, multi-view clustering~\cite{Bickel2004}, co-clustering~\cite{dhillon2001_co-clustering,dhillon2003_information}, biclustering~\cite{mirkin1996_clustering}, two-mode clustering~\cite{Mechelen2004_two_mode_clustering}, direct clustering~\cite{Hartigan1972_direct_clustering}, and block clustering~\cite{govaert2008_block_clustering}, by preserving the similarity between samples of the same concept in each \emph{description space}, i.e., when observing the \emph{a priori} defined subsets of features in isolation~\cite{Lanfermann2022b}.
Such a description space comprises of a subset of features that characterizes the instances in one aspect~\cite{Lanfermann2022b}.
A definition of the key terms used in this work can be obtained from Table~\ref{tab:terms}.

\begin{table*}[!ht]
\centering
\caption{Description of key terms
}
\begin{tabularx}{0.98\textwidth}{lX}  
\toprule
Term & Description \\
\midrule
Feature & A property that is attributed to a data sample (e.g., investment cost) \\
Data sample & A datum or an instance of data, given as a set of features that are attributed to the instance\\
Data set & A set of data samples \\
Description space & The space that is spanned by a group of features (e.g., the combination of investment cost and emissions)\\
Concept & A set of data samples that is similar in more than one description space\\
Concept quality metric & A numerical quality measure for one set of concepts \\
Concept identification & The association of data samples into concepts by optimizing a concept quality metric\\
\bottomrule
\end{tabularx}
\label{tab:terms}
\end{table*}

The concept identification method employed in existing work~\cite{Lanfermann2020,Lanfermann2022a} may be viewed as a special form of clustering technique~\cite{Lanfermann2022b}, that differs from existing methods as it aims at uncovering clusters of samples which are non-overlapping and consistent with respect to multiple description spaces of the joint feature space~\cite{Lanfermann2022a}. 
Consistency is given for a cluster, if the instances that are associated with the cluster are assigned to the same cluster in arbitrary, \textit{a priori} defined description spaces.
Concepts are then defined as non-overlapping, consistent clusters of solutions in the set of all solutions. 
This developed methodology for concept identification in multiple description spaces can steer the identification process towards a consistent and meaningful distribution of concepts. 

However, the \textit{a priori} determination of suitable description spaces has not yet been addressed in previous work, although it has a significant influence on the outcome.
Previous studies \cite{Lanfermann2022a} show that allocating features into different description spaces introduces a correlation of those features for the identifiable concepts, while having features in the same space allows for arbitrary combinations of those.
Depending on the requirements of a given identification task, both those properties can be either beneficial or unwanted.
In any case, only a thorough selection and partitioning of features into description spaces can lead to plausible and useful concepts with respect to the requirements and provide helpful insights for a decision maker.

Motivated by this, in this work, we propose to thoroughly investigate the impact of the choice of description spaces on the identified concepts in the energy management domain.
It should be noted that to the best of our knowledge, the existing state of the art clustering methods cannot be readily employed to solve the problem of identifying consistent groups of data that are similar with respect to multiple pre-defined subspaces of the data set. 
Therefore, direct a comparison of our results to other state of the art methods is not possible in this work.

\section{Investigation of the impact of the choice of description spaces in concept identification}\label{sec:choice_of_description_spaces}

Motivated by the benefits of concepts in analyzing optimal solutions, this work propose applying the concept identification approach to a many-objective data set from the energy management domain. This work first focuses on thoroughly investigating the impact of the choice of description spaces on the identified concepts and the effect of allocating features of the data set into separate description spaces.
Then, after identifying concepts, this work is expected to provide valuable insights and guide the decision maker towards finding suitable options for efficient energy management configurations which can be aligned with the decision maker's expectations and constraints. 

\begin{figure*}[tbp]
\centering
\includegraphics[width=1.0\textwidth]{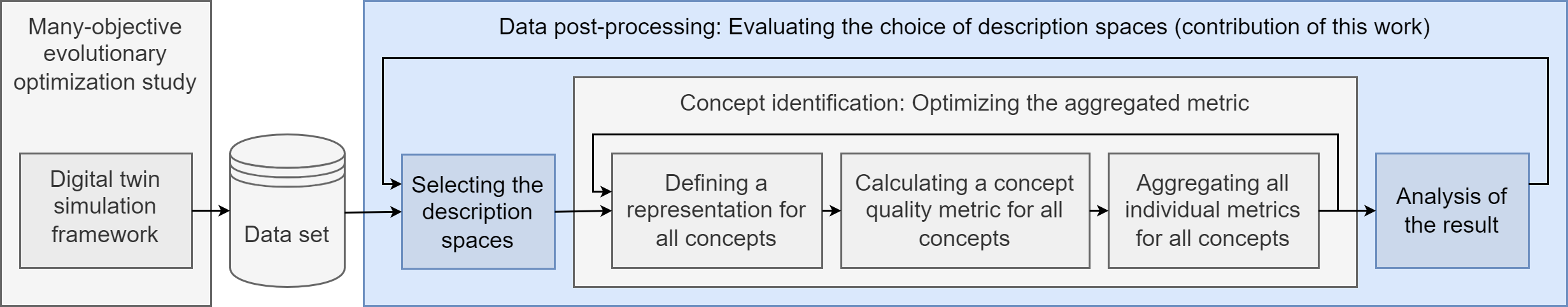}
\caption{
Process of evaluating the choice of description spaces: A data set, obtained from a many-objective evolutionary optimization study in previous work~\cite{Liu2022}, is investigated for concepts.
The additional step of selecting different description spaces is added in an iterative way to the standard concept identification~\cite{Lanfermann2022a} procedure.
}
\label{fig:ci_flowchart}
\end{figure*}

In concept identification, the aim is to group solutions into different concepts based on selected description spaces, and the choice of description spaces can strongly impact the concepts which are identified.
Thus, for the experiments in this work, we add an additional iterative step to the concept identification approach, with which the selection of description spaces is investigated (Fig.~\ref{fig:ci_flowchart}).
For several different sets of description spaces, we conduct a concept identification process and compare the outcome.

Algorithmically, the concept identification approach is implemented as a procedure, where 
a concept quality metric $Q$ is maximized in an evolutionary optimization framework.
The optimization adapts the representation of each concept in each description space and arrives at an optimized distribution of concepts. 
The concept identification approach targets to create concepts which are as large as possible, but which do not have any overlap.
Ideally, all data samples belong to exactly one concept. 
The quality metric also takes into account that concepts should be neither too small nor too large, and allows the user to  provide preference samples which should be part of a concept.  
The exact formula to calculate the quality metric for the concepts is not reproduced here, and the interested reader is referred to the original literature~\cite{Lanfermann2022a}.

\begin{figure*}[tbp]
\centering
\includegraphics[width=1.0\textwidth]{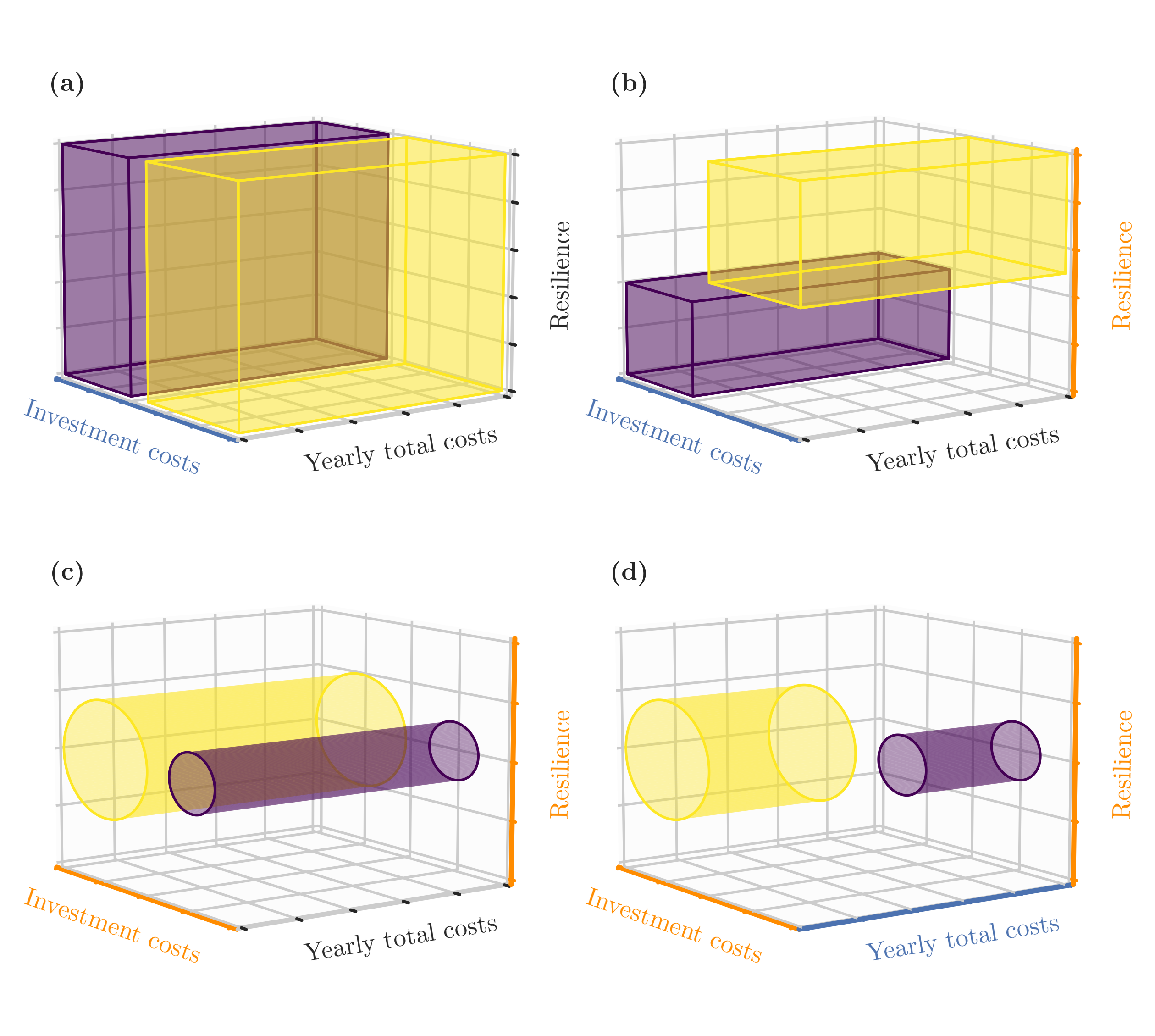}
\caption{
Several alternative choices for selecting the description spaces for an example with three features.
The selected features and their associated description space are indicated by the color of the axis labels. 
Black labels indicates that the corresponding feature is not associated with any description space.
The colored boxes indicate possible concepts for the different choices of description spaces for a case where two concepts should be identified. 
    (a) A single one-dimensional description space (Investment cost in blue). 
    (b) Two separate one-dimensional description spaces (Investment cost in blue, Resilience in orange)
    (c) A single two-dimensional description space (Investment cost and Resilience in orange)
    (d) One two-dimensional and one one-dimensional description space (Investment cost and  Resilience in orange, Yearly total cost in blue).
    }\label{fig:guidelines}
\end{figure*}

In order to show the influence of the choice of description spaces in a preliminary example, we consider a data set containing data samples that each can be described by three features (investment cost, yearly total cost and resilience) and we illustrate the possible four choices in the following, as shown in Fig.~\ref{fig:guidelines}.
\begin{itemize}
    \item If only one feature is considered as the only description space, the potential concepts will be discriminated based on the one single feature alone.
    If, for example, two concepts were to be identified, the three-dimensional space would be divided into two boxes that each could \emph{host} one concept (Fig.~\ref{fig:guidelines} (a)). 
    Samples can then be assigned to one of the two different concepts based on the value of the one chosen feature alone.
    \item If, however, two features are chosen as separate description spaces, the potential concepts would be divided based on both of these features separately.
    A concept identification process that aims at identifying two concepts would therefore divide the three-dimensional space into two non-neighboring boxes alongside the chosen features (purple and yellow boxes in Fig.~\ref{fig:guidelines} (b)), or analogously the empty space).
    The two concepts would then be located in those boxes.
    The requirement that the identified concepts must not overlap even when only one description space is viewed, i.e., when the data is projected onto one of the chosen feature axis in this example, leads to this restriction to non-adjacent regions.  
    \item If a combination of two features were chosen to span one description space, the potential concepts will be non-overlapping regions in the two-dimensional plane spanned by those two features.
    The concept regions defined in the two-dimensional description space are then extruded in the dimension of the remaining feature.
    Choosing the shape of the concept regions as ellipses in the description space, the two regions available to the concepts are given as two cylindrical volumes (Fig.~\ref{fig:guidelines} (c)). (Other arbitrary shapes can be applicable for the concept regions, depending on the preference of the decision maker.)
    Comparing this choice with one description space containing both two features to the previous setup (as in Fig.~\ref{fig:guidelines} (b)) where both features are put into separate description spaces elucidates one core aspect of this concept identification approach, that is, concepts are non-overlapping even if several features are projected onto a single description space. 
    Fig.~\ref{fig:guidelines} (b) implies that each feature value can be separately used to uniquely identify a concept. 
    In contrast, in Fig.~\ref{fig:guidelines} (c), both feature values are necessary to uniquely identify a concept.
    In general, the concepts will overlap if they are projected onto only one feature (resilience in the example).    
    \item If two separate description spaces were chosen, one containing two features and the other one feature, the potential concept regions for two concepts partition the full three dimensional feature space into four regions.
    Assuming the same ellipsoidal concept regions in the two-dimensional description space as before, the extruded cylindrical volumes would be separated into two disjoint parts along the one-dimensional description space.
    The two concepts could then be identified in two non-neighboring volumes (yellow and purple in Fig.~\ref{fig:guidelines} (d)). 
    Other locations of the concepts are forbidden due to the requirement of non-overlapping concepts in  the projections into each  description space.  
\end{itemize}

It should be noted that the above described regions only characterize the possible locations for concepts, and data samples located in one of these regions are not automatically associated with a concept. 
For complex data samples, there are many data samples inside these regions that are not associated with any concept. 
Also, as indicated in the examples above, the description spaces used for identifying the concepts do not need to include all features. 

These illustrative example sketches should make it clear that the choice of description spaces has a large influence on the identifiable concepts.  
However, a reasonable selection of description spaces is often not intuitive and a difficult task on its own. It is believed that the choice of description spaces should align with the preferences of the decision maker in order to provide the most useful concepts.

The decision maker has to decide whether each feature should be considered in the concept identification approach, and if so, in which description space it should be included. 
Two essential insights from the previous discussion can be gained:
(i) Features in the same description space can be arbitrarily combined in one concept and every possible combination of feature values can also be represented by a separate concept.
(ii) Features in different description spaces lead to stronger correlations in the feature values for concepts. 
Because the feature values in one description space condition the feature values in another description space, only a subset of all possible combinations of feature values can be represented in concepts. 

So, for example, if a set of samples needs to be uniquely identifiable on the basis of one feature value, this feature needs to be considered as a separate one-dimensional description space.
Similarly, if a set of samples needs to be uniquely identifiable based on the combination of multiple features, all these features need to be considered in one description space. 
On the other hand, if features are assigned to different description spaces, the feature values in one description space impose a condition on the feature values in the other description spaces. 
Of course, how strong this conditioning and the induced correlations will be, strongly depends on the structure of the data set. 
But in any case, the resulting concepts can only represent subsets  of combinations of these feature values.

\section{Identification of configuration concepts for building energy management}\label{sec:results}

\subsection{Description of the energy management configuration problem}\label{subsec:configuration_problem}

The data set under investigation has been created in the context of a study on many-objective evolutionary optimization algorithms~\cite{Liu2022}.
It has been created with a realistic high-quality Digital Twin simulator which was calibrated on real measurement data from a company facility.
The target has been to find a large set of Pareto-optimal configurations by changing the parameters of power supply components, such as \acrshort{pv} system, battery storage, and heat storage.
Each configuration is defined by nine different parameters which are listed in the configuration Table~\ref{tab:parameters}.

\begin{figure*}[tbh]
\centering
\begin{tikzpicture}


\node[inner sep=0pt, label={[align=center]south:External Grid Connection}] (exgrid) at (1,1.5)
    {\includegraphics[width=.1\textwidth]{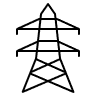}};

\node[inner sep=0pt, label={[align=center]south:Weather Data}] (weather) at (1,6)
    {\includegraphics[width=.1\textwidth]{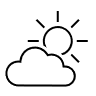}};

\node[inner sep=0pt, label={[align=center]south:PV System}] (pv) at (3,4)
    {\includegraphics[width=.1\textwidth]{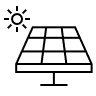}};

\node[inner sep=0pt, label={[align=center]south:Building Power Consumption}] (building) at (7,4)
    {\includegraphics[width=.1\textwidth]{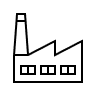}};

\node[inner sep=0pt, label={[align=center]south:Combined Heat and Power Plant}] (chp) at (12,4)
    {\includegraphics[width=.1\textwidth]{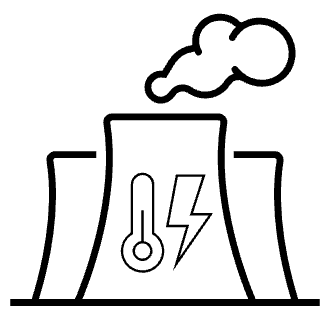}};
    
\node[inner sep=0pt, label={[align=center]east:Battery Storage And Controller}] (battery) at (7,0.5)
    {\includegraphics[width=.1\textwidth]{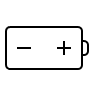}};

\draw [thick, >=latex, ->] (weather) -| (pv);  
\draw [thick, >=latex, ->] (weather) -| (building);  
\draw [thick, >=latex, ->] (weather) -| (chp);  

\draw [thick, >=latex, ->] (pv) -- (building);  

\draw [thick, >=latex, <->] (building) -- (chp);  
\draw [thick, >=latex, <->] ([yshift=+0.4cm]battery.center) -- ([yshift=-1.7cm]building.center);

\draw [thick, >=latex, <->] (exgrid) -| ([xshift=-0.3cm, yshift=-1.7cm]building.center);

\end{tikzpicture}
\caption[Energy management system]{
Simulation model of the energy management system: The model simulates the building power and heat demand based on time and weather conditions.
Energy is provided by the grid connection, a \acrshort{pv} system, a combined heat and power plant (\acrshort{chp}) and a stationary battery.
The battery's charging and discharging behavior is controlled depending on a predetermined control strategy and internal reference values, e.g., the overall power consumption level.
}
\label{fig:energy_system}
\end{figure*}

For each configuration a set of performance values, i.e., objectives, are evaluated using the Digital Twin simulation model of an existing research campus building~(Fig.~\ref{fig:energy_system}).
The simulations are performed with the commercial tool SimulationX\footnote{\url{https://www.esi-group.com/products/system-simulation}}, which is based on the Modelica simulation language\footnote{\url{https://modelica.org/modelicalanguage.html}} and uses the realistic Green City Library\footnote{\url{https://www.ea-energie.de/en/projects/green-city-for-simulationx-2/}}.
Individual components as well as a small set of configurations of the Digital Twin simulation model have been calibrated with real-world measurement data~\cite{schwanDigitalTwin2019}   
From the simulation results, a set of ten different partially related and generally conflicting objectives are calculated, which are relevant to produce an informed investment decision. 
These objectives are listed in Table~\ref{tab:objectives}.
The \textit{Investment cost} are directly determined by the cost for the battery, the PV and the heating storage systems. 
\textit{Annual operation costs} due to maintaining and operating the whole system will involve the grid electric cost, the gas consumption, the peak electricity load cost and the CHP maintenance cost.
The objective \textit{Resilience} is defined as the duration the company would be able to operate in case no grid power is available, thus if energy is only provided by  the local production \acrshort{pv} system, combined heat and power (\acrshort{chp}) and from battery storage.
\textit{Medium SOC time share} refers to the inverted time ratio in which the battery state of charge (\acrshort{soc}) of the battery resides between \SI{30}{\percent} and \SI{70}{\percent}, which estimates aging of the battery.
More details on all objectives and parameters can be found in the original work~\cite{Liu2022}.

In \cite{Liu2022}, state-of-the-art surrogate-assisted evolutionary algorithms such as 
RVMM~\cite{rvmm}, K-RVEA~\cite{chugh2016surrogate}, REMO~\cite{Hao2022}, 
are utilized to produce a large data set of high quality in handling the many-objective energy management problem.

From the entirety of all feasible solutions that these algorithms produce, all dominated solutions are discarded and only 
\SI{20000}{} non-dominated Pareto-optimal configurations are used for further analysis in this work.
More details on the parameters, objectives and the multi-objective optimization approaches  can be obtained from~\cite{Liu2022} while the simulation approach itself is discussed in~\cite{Unger2012}.

\begin{table*}[h]
\centering
\caption{Design parameters of the energy management configuration problem. The hyphen symbol indicates that no unit is associated with the corresponding parameter.}
\begin{tabularx}{\textwidth}{m{0.5in} m{3.5in} m{0.5in} m{0.5in} l}
\toprule
& Parameters & Min & Max & Unit\\
\midrule
$\alpha_{PV}$   & Inclination angle of PV system    & 0     & 45    & \degree \\
$\beta_{PV}$    & Orientation angle of PV system    & 0     & 360   & \degree \\
$P_{PV}$        & Peak power of PV system           & 10    & 450   & kW \\
$C_b$           & Nominal Battery Capacity          & 5     & 1000  & kWh \\
$b_\mathrm{max}$       & Maximum Battery state of charge   & 0.50  & 0.95  & - \\
$b_\mathrm{min}$       & Minimum Battery state of charge   & 0.05  & 0.40  & - \\ 
$P_{c}$         & Battery Charging Threshold        & -500  & 149.9 & kW \\ 
$P_{d}$         & Battery Discharging Threshold     & 150   & 700   & kW \\ 
$V$             & Heat storage cylinder volume      & 1     & 5     & m\textsuperscript{3} \\
\bottomrule
\end{tabularx}
\label{tab:parameters}
\end{table*}

\begin{table*}[h]
\centering
\caption{Objectives of the energy management configuration problem. The hyphen symbol indicates that no unit is associated with the corresponding objective, or that there is no upper or lower bound.}
\begin{tabularx}{\textwidth}{m{0.5in} m{4.7in} m{1in}} 
\toprule
& Objectives & Unit \\
\midrule
$C_\text{invest}$    & Investment cost                               & Euro \\
$C_\text{annual}$    & Annual operation cost                        & Euro \\
$R$             & Resilience                                    & s \\
$G$             & Yearly CO\textsubscript{2} emissions                     & t \\
$\bar{b}$       & Mean battery state of charge                  & - \\
$E_{d}$         & Yearly discharged energy (from the battery)   & kWh \\
$P_{p}$         & Maximum power peak                            & kW \\
$t$             & Medium \acrshort{soc} time share                         & - \\
$E_{f}$         & Yearly feed-in energy                         & kWh \\
$P_{f}$         & Maximum feed-in power peak                    & kW \\
\bottomrule
\end{tabularx}
\label{tab:objectives}
\end{table*}

\subsection{Experimental results}

The analysis of the trade-off relations between objectives becomes more difficult with the increase of the number of objectives. Since the optimization of the building energy management involves ten objectives, it is very challenging to assess the trade-off relations between all those objectives and arrive at an informed decision for selecting the most appropriate configurations given some specific design goals. 
Therefore, it is highly desirable to provide a reasonable selection approach to a decision maker to allow for an educated investment decision for the ten-objective building energy management problem.
Possible solution candidates need to fulfill certain economic requirements, and span the trade-off options along relevant criteria which are  predefined by the decision maker.

Based on the \SI{20000}{} Pareto-optimal configuration options, a series of experiments is conducted to illustrate how energy management configuration concepts can be identified from the present data set in a meaningful way.
In the first experiment, only three major objectives related to cost end resilience are considered in the identification process to highlight the universal importance of the choice of description spaces.
In the second experiment, two parameters and six objectives are selected and split into description spaces to identify technically feasible and useful concepts.
In the third experiment, the samples from one of the concepts in the second experiment are re-analyzed for a further in-depth assessment of the concept options based on the three description spaces obtained in the first experiment.

The concept identification process is conducted by first specifying the number of concepts to be identified. 
Each potential concept region is given by a parameterized geometric shape, where \mbox{(hyper-)ellipses} are used. 
The parameters of the concept regions are obtained by optimizing a metric characterizing the quality of the resulting concepts using an evolutionary optimization approach (We use CMA-ES\cite{hansenCMA} in this work). 
The optimization is run for 1000 generations with a population of $\lambda=20$ candidate solutions. 
The details of the concept identification approach are described in \cite{Lanfermann2022a}.

\subsection{Experiment 1: Concept identification based on three main objectives}\label{subsec:experiment_1}

An investment decision for an energy management building configuration has to be made, given  multiple major external requirements and constraints.
Key aspects are initial investment cost, the annually recurring cost for maintenance, and resilience as measures for the effectiveness of the installations (see Section \ref{subsec:configuration_problem}).
A reasonable request to the engineer would be to deliver configuration concepts that are distinguishable in terms of investment cost, while also providing trade-off options between maintenance cost and resilience.
That means the identified concepts should not be ranked based on their investment cost alone.
This is achieved by splitting the corresponding objectives into two separate description spaces, the first consisting of the investment cost only, and the second comprising yearly cost and resilience together.

\begin{figure*} 
\textbf{(a)}\\[2mm]
\includegraphics[width=\textwidth]{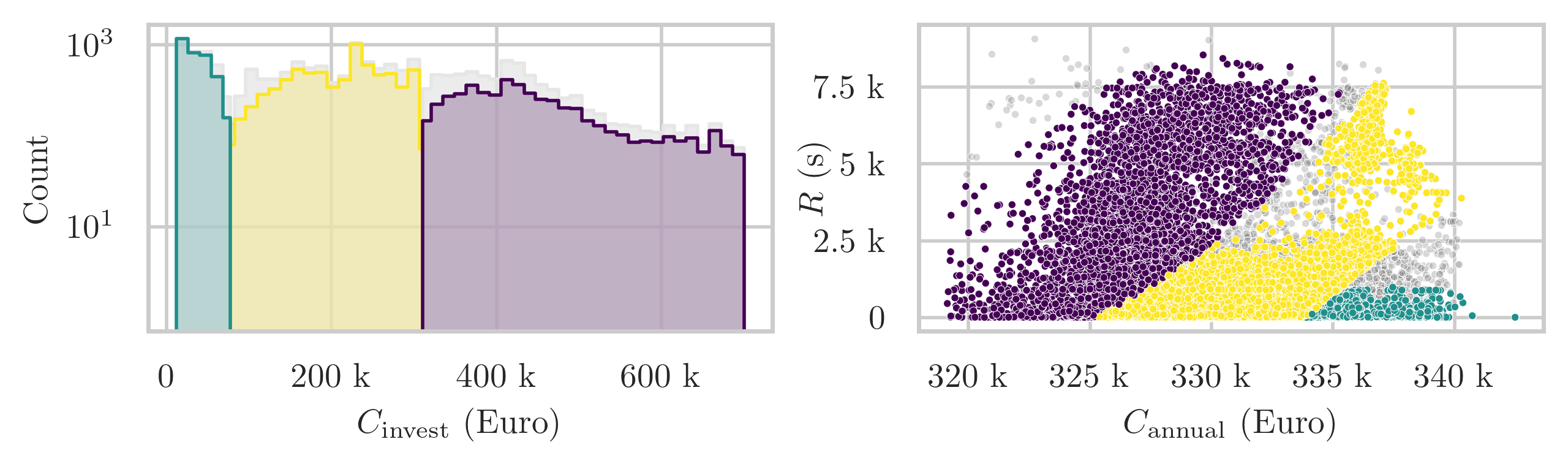}\\
\textbf{(b)}\\[2mm]
\includegraphics[width=\textwidth]{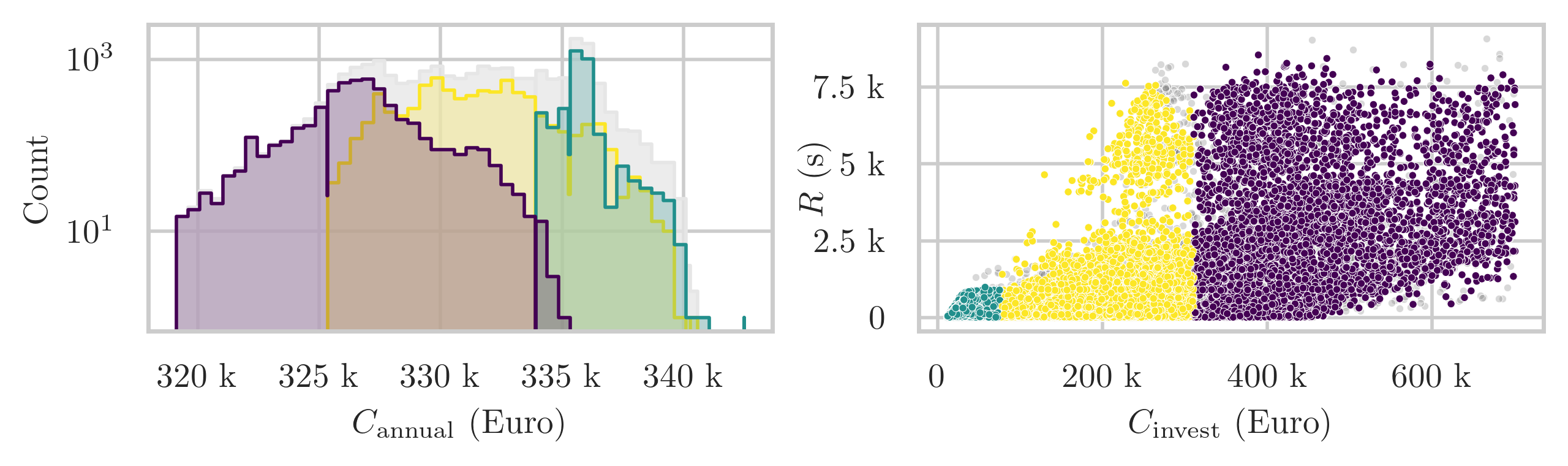}\\
\textbf{(c)}\\[2mm]
\includegraphics[width=\textwidth]{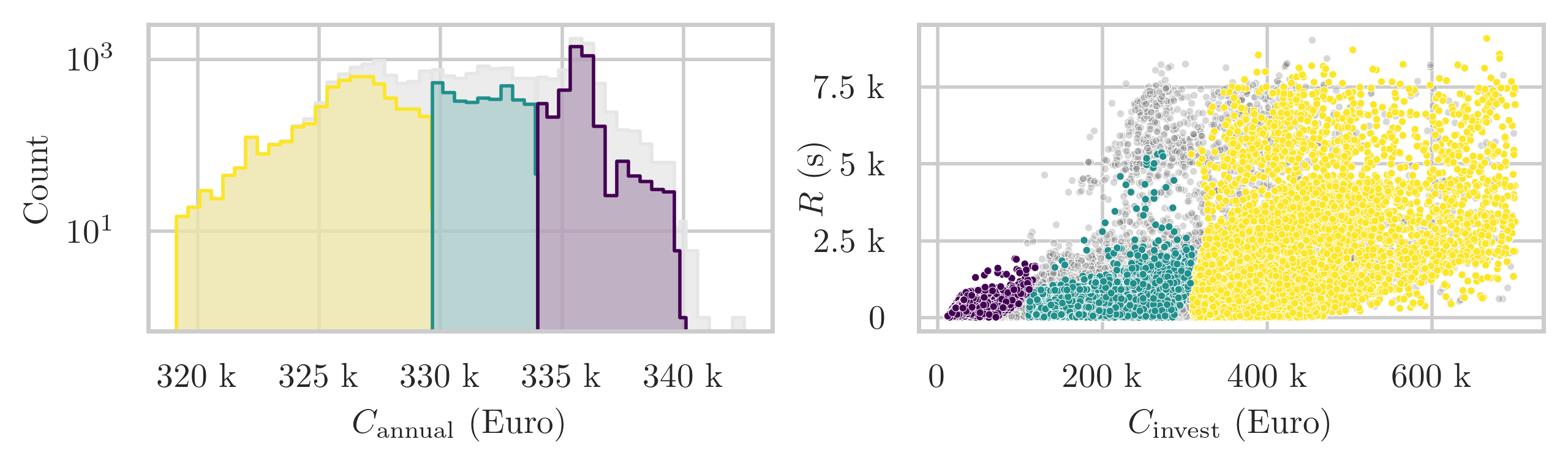}
\caption{
(a) Identified concepts from experiment 1: The process identifies three concepts that satisfy the given requirements imposed by the two description spaces. 
(b) The same concepts are not reasonable for a different combination of description spaces. 
(c) A separate concept identification process for the second distribution of description spaces also leads to reasonable concepts.
}
\label{fig:DSprojection}
\end{figure*}

The target for the concept identification process is to identify three concepts in these two description spaces. 
As required, the three identified concepts represent high, medium, and low investment cost solutions, as can be seen by the purple, yellow, and green concepts in the left panel of Fig.~\ref{fig:DSprojection}~{(a)}. 
In the second description space (right panel of Fig.~\ref{fig:DSprojection}~{(a)}), the three concepts encompass trade-off options between yearly cost and resilience values.
The three concepts fully meet the requirement for the decision process. 
The concept that includes the configurations with the highest investment cost (purple) indeed includes the configurations with the best trade-off between the largest resilience and the lowest annual cost. 
On the other hand, the concept encompassing the lowest investment cost configurations (green) leads to the overall largest annual cost with rather low resilience. 
A typical situation of a complex real-world data set can be observed as well, since many samples are not associated with any concept at all (grey samples). 
This is prominently visible in the second description space (right panel) as many samples with larger resilience, which are located above the low investment cost concept (green samples), are not associated with any concept. 
This is a direct consequence of choosing the investment cost as a separate description space, as these solutions have too large investment cost and would be overlapping with the yellow concept in this description space.  

To illustrate how the choice of the description spaces affects the concepts, a different setup is considered in the following. 
The first description space is given by the annual cost alone, while the second space is spanned by investment cost and resilience.
The economic reason for such a choice is different, as it puts most emphasis on the annual cost, along which all concepts should be ranked, while the trade-off between the other two objectives is considered. 
In Fig.~\ref{fig:DSprojection}~(b), the concepts identified for the previous choice of description spaces are plotted in the novel setup.  
It can be observed that in the new projection, the previously identified concepts do not meet the requirements of the decision maker. 
The large visible overlap between all three groups in the first description space  prohibits a unique association of configurations to concepts based on annual cost alone.
Also, the second description space clearly shows the separation along the investment cost (as imposed by the previous concept identification process), but does not provide trade-off options regarding investment cost and resilience to the decision maker. 

However, conducting the concept identification process within this new set of description spaces gives the desired results (Fig.~\ref{fig:DSprojection}~(c)). 
The process delivers three unique concepts, separable with respect to the isolated feature of yearly total cost, while at the same time providing trade-off options for the joint space of investment cost and resilience.

The first experiment demonstrates that the developed concept identification approach can be used to define meaningful concepts in complex data sets and provides valuable insights as basis for an informed decision. 
The freedom to choose the partitioning of the full feature space into description spaces allows to meet the requirements of the  decision maker, since the potentially identifiable concepts are significantly impacted by the definition of the description spaces.

\subsection{Experiment 2: Concept identification based on nine parameters and six objectives}\label{subsec:experiment_2}

The previous example only considers the three major objectives related to cost and resilience, while ignoring all other features. 
To make a well-informed decision, more aspects of the configurations, and in particular some technical aspects should be considered as well.  
To assure the technical feasibility and economic sensibility of selected samples, two parameter values and six objectives are involved in making a concept identification, and they are split into four distinct description spaces, as presented in Table~\ref{tab:ex2}.

\begin{table*}[]
\caption{List of description spaces and respective variables for all experiments.
}
\label{tab:ex2}
\begin{center}
\begin{tabular}{ l c c c c}
    \toprule
    & Description space 1 & Description space 2 & Description space 3 & Description space 4 \\ [0.5ex] 
    \midrule
    Exp. 1a, 3    & $C_\text{invest}$ & $C_\text{annual}$, $R$        &                               & \\ [0.5ex] 
    Exp. 1b, 1c   & $C_\text{annual}$ & $C_\text{invest}$, $R$        &                               & \\ [0.5ex] 
    Exp. 2        & $P_{PV}$, $C_b$   & $C_\text{invest}$, $P_{p}$    & $C_\text{annual}$, $E_{d}$    & $\bar{b}$, $E_{f}$ \\ [0.5ex] 
    \bottomrule
\end{tabular}
\end{center}
\end{table*}

The two parameters ($C_b$ and $P_{PV}$), are chosen to be represented in the same description space (Description space 1 in Table~\ref{tab:ex2}), as any combination of these values can make sense and can create a valid configuration.
The six objectives are separated into three description spaces (Description space 2 to 4 in Table~\ref{tab:ex2}), each of which considers only two objectives.
This choice is motivated by the following insights.
The cost variables $C_\text{invest}$ and $C_\text{annual}$ are placed into two separate description spaces to allow for distinction between concepts based on either of these two values.
This helps to avoid uninformative concepts where, for example, the investment cost is high but the annual cost ranges across all possible values.
The objectives $\bar{b}$ (mean battery state of charge) and $E_{f}$  (energy fed into the grid) are represented in the same description space, as combinations of these two values naturally map to operating modes such as cost reduction via peak-shaving (low $E_f$ and large $\bar b$) or cost reduction via \acrshort{pv} utilization (large $E_f$).
Similar to the cost objectives, $P_{p}$ and $E_{d}$ are separated to avoid unwanted concepts with high grid supply power peaks and large amounts of discharged energy.
This should make sure that if an expensive battery energy storage system is implemented, it will be used for power peak shaving.
$E_{d}$, $\bar{b}$, and $C_\text{invest}$ are also separated to avoid unwanted concepts that represent configurations with large energy capacity (high $C_\text{invest}$), that is not used efficiently (high $\bar{b}$ and low $E_{d}$).

It should be noted that these choices of description spaces are not obligatory to achieve the stated goals. 
They should rather serve as guidelines, and especially for complex situations where many features need to be included, some choices might need to be reconsidered after inspecting various outcomes of the concept identification process.

For this example, the concept identification approach also includes 30 solutions of particular interest in the process as user preference.
These are specified by the decision maker \textit{a priori} and the concept identification process then generates concepts which should include these samples. 
Thus, these user preference samples are a means to anchor the concepts in various regions of interest. 
For the current example these are:  

\begin{itemize}
    \item Ten samples with very low investment cost
    \item Ten samples with good trade-offs between low investment cost and a low power peak
    \item Ten samples with low annual cost
\end{itemize}

\begin{figure*}[tbph]
\centering
\includegraphics[width=1.0\textwidth]{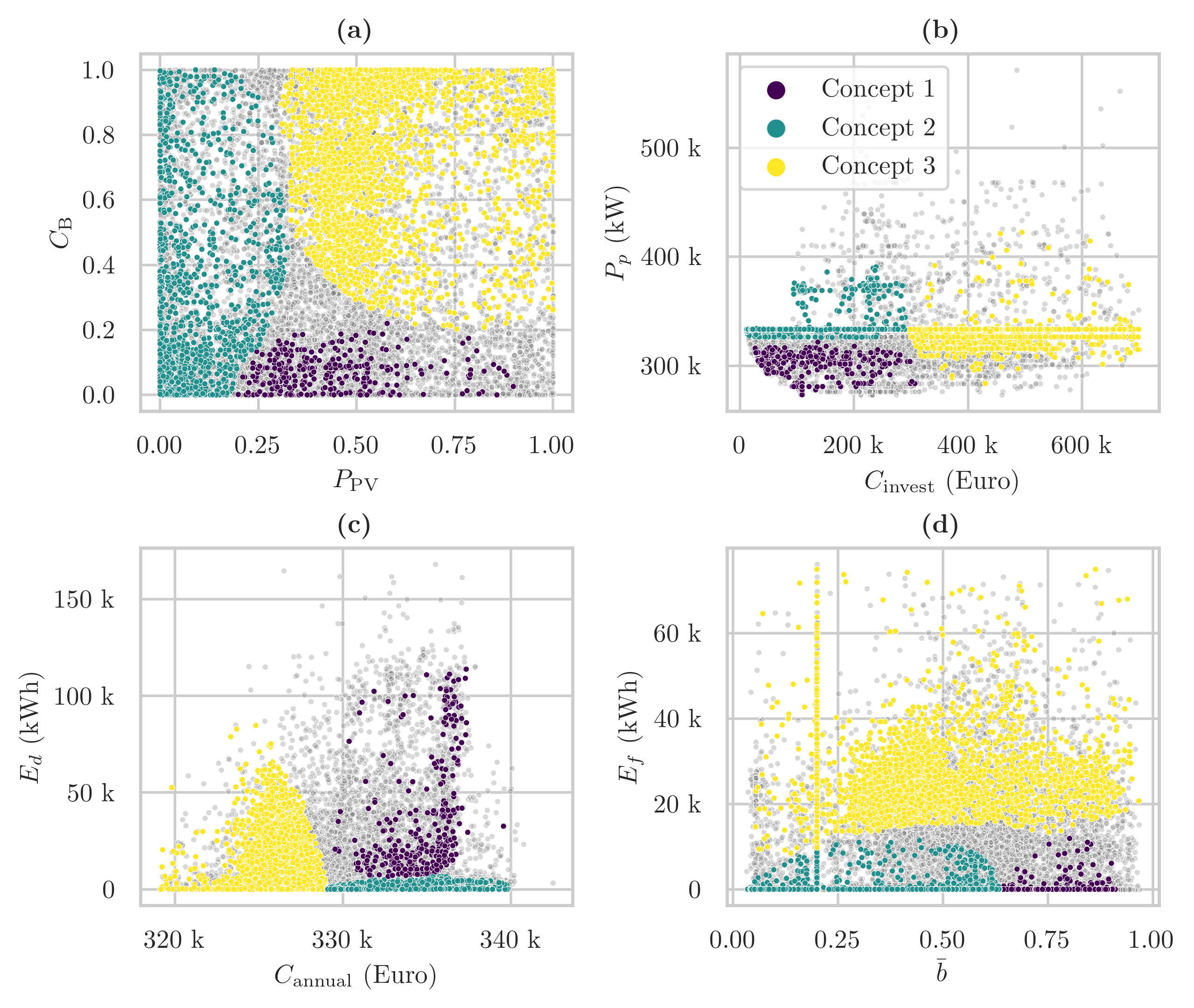}
\caption{Identified configuration concepts based on four description spaces: (a) $P_{PV}$ and $C_b$, (b) $C_\text{invest}$ and $P_{p}$, (c) $C_\text{annual}$ and $E_{d}$, (d) $\bar{b}$ and $E_{f}$.
The colored samples (purple, green, and yellow dots) are associated with concepts 1, 2, and 3, respectively. 
The grey samples are not associated with any concept.
The two parameters of description space 1 (top left plot) are normalized.
} 
\label{fig:core_samples_09}
\end{figure*}

\begin{figure*}[tbph]
\centering
\includegraphics[width=1.0\textwidth]{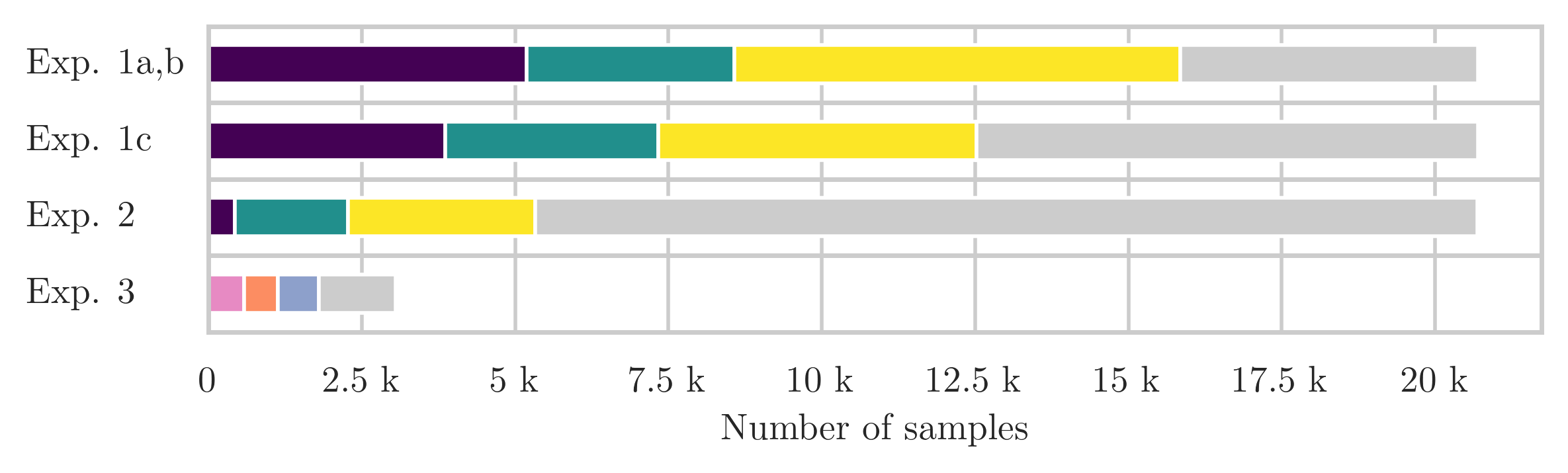}
\caption{Number of samples for each concept for all experiments: Portion of samples that are allocated to one of the identified concepts (colored proportions) and samples that are not associated with a concept (grey).
} \label{fig:concept_sizes}
\end{figure*}

In this example the concept identification process is conducted by optimizing the concept quality metric for three concepts using again a CMA-ES for 370 generations and a population size of 61.
The process identifies three concepts (Fig.~\ref{fig:core_samples_09})
which contain 435 (purple), 1845 (green), and 3054 (yellow) samples (see Exp. 2 in Fig.~\ref{fig:concept_sizes}).
The remaining samples are not associated with any concept.

In the first description space of the objectives (top right of Fig.~\ref{fig:core_samples_09}), a primary division based on investment cost is visible.
Concept 1 (purple) and concept 2 (green) are associated with low investment cost, while concept 3 (yellow) shows high investment cost.
An intuitive reciprocal trend is seen in objective space 2 (lower left): concept 3 is associated with low annual cost, while concepts 1 and 2 show high annual cost.
The size of the \acrshort{pv} system and the battery are two factors that have a high impact on the investment cost.
Consequently, concept 3 has both, large \acrshort{pv} systems and large batteries, while concept 1 utilises small batteries but larger \acrshort{pv} systems ($P_{PV}$) and concept 2 includes small \acrshort{pv} systems but large batteries ($C_b$).
The configurations of concept 3 lead to a high amount of \acrshort{pv} produced electricity, which is fed back into the grid ($E_f$) to a significantly larger extent than for concept 1 and 2.
Accordingly, this assures overall lower annual cost.

A secondary division between concepts 1 and 2 is present along the objectives maximum power peak ($P_{p}$), yearly discharged battery energy ($E_{d}$), and mean battery \acrshort{soc} ($\bar{b}$).
Concept 1 has generally lower power peak values, higher amounts of energy discharged from the battery and a higher average battery \acrshort{soc} than concept 2.
Concept 1 therefore represents configurations where power peak shaving is done resulting in low $P_p$, which requires a battery with relatively high mean \acrshort{soc} to be readily used as soon as a imminent power peak is detected. 
Concept 2 represents low investment cost solutions containing only small \acrshort{pv} systems. 
However, it includes configurations with batteries of all sizes, but where the battery is not effectively used in general. 
Many configurations utilize large batteries without specific benefit to the overall system. 
Therefore, this concept does not represent one coherent set of configurations, and further analysis is necessary to distinguish useful configurations. 
It would be sensible to refine concept 2 in a subsequent post-processing step, e.g., to find the best solutions with small \acrshort{pv} systems and small batteries within the concept.

\begin{figure*}[h]
\centering
\includegraphics[width=1.0\textwidth]{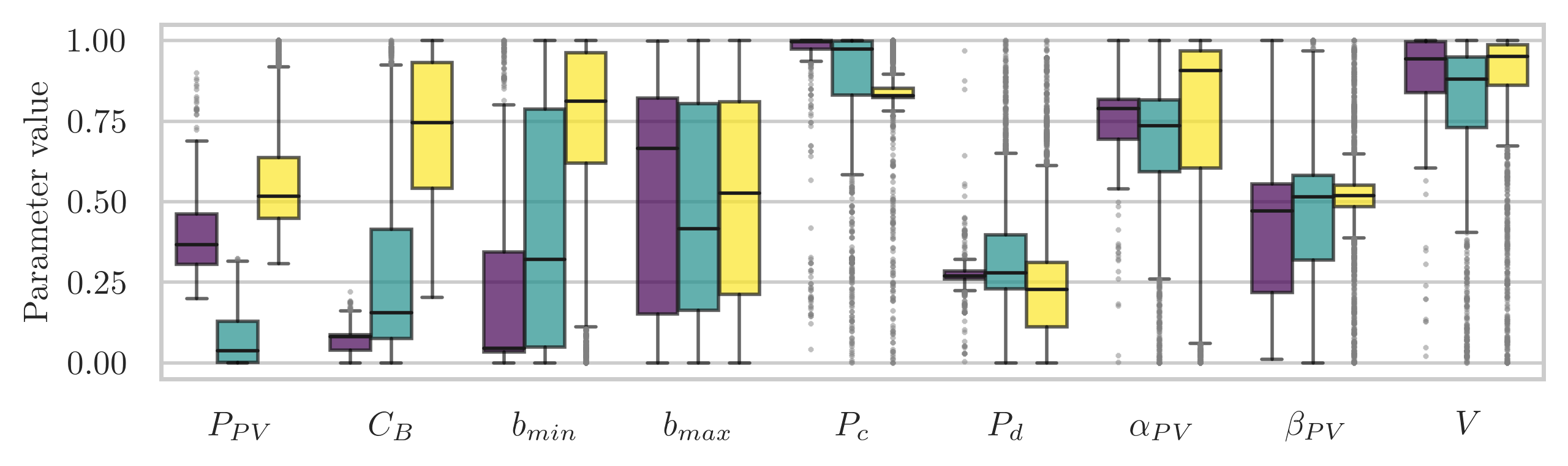}
\caption{Distribution of parameters for all identified concepts 1 (purple), 2 (green), and 3 (yellow)
} \label{fig:ex2_parameters}
\end{figure*}

Further insights into the concepts can be gained by analyzing their parameter values (Fig.~\ref{fig:ex2_parameters}). 
While the concepts are not different with respect to some parameters like maximum battery \acrshort{soc} $b_\mathrm{max}$, \acrshort{pv} inclination and orientation angles $\alpha_{PV}$ and $\beta_{PV}$,  others clearly reveal systematic differences.   
For example, the battery controller parameters $b_\mathrm{min}$ (minimum battery SOC) and $P_c$ (battery charging thresholds) have characteristically different ranges for the concepts.
In particular, they confirm the difference in the battery utilization between the two low-investment cost concepts 1 (purple) and 2 (green). 
On average, the solutions in concept 1 have lower minimum battery SOCs ($b_\mathrm{min}$) and higher battery charging thresholds ($P_c$) than concept 2.
This results in a behavior of solutions in concept 1, where the batteries are charged even when the building is consuming energy from the grid, and the batteries can be almost completely discharged when necessary.

Concept 3 represents the opposite to concept 1: generally higher investment, but low annual cost, with large \acrshort{pv} systems and large batteries.
However, it covers a relatively large span of the objective values, for example in terms of investment cost. 
For illustrative purposes, it is assumed that the high-investment cost segment is of interest to the decision maker, and consequently, concept 3 is chosen to be refined in a secondary concept identification step, which will be discussed in detail in the following subsection.

\subsection{Experiment 3: Concept identification based on three objectives for high investment cost solutions}\label{subsec:experiment_3}
In this subsection, only the samples from concept 3 of the previous experiment 2 are selected for further refinement.
Investment cost is chosen as one description space, resilience and annual cost as another, which is the same setup as experiment 1 in Section~\ref{subsec:experiment_1}.
This way, technically meaningful configurations with high investment cost of the previous analysis can be re-analyzed.
In terms of the trade-off between annual cost and resilience, they can be segmented according to investment cost again.

The concept identification process is again conducted by optimizing the concept quality metric using evolutionary optimization (CMA-ES, 400 generations, population with 22 candidates).
The algorithm is configured to identify three concepts, and no preference samples are specified.

\begin{figure*}[h]
\centering
\includegraphics[width=1.0\textwidth]{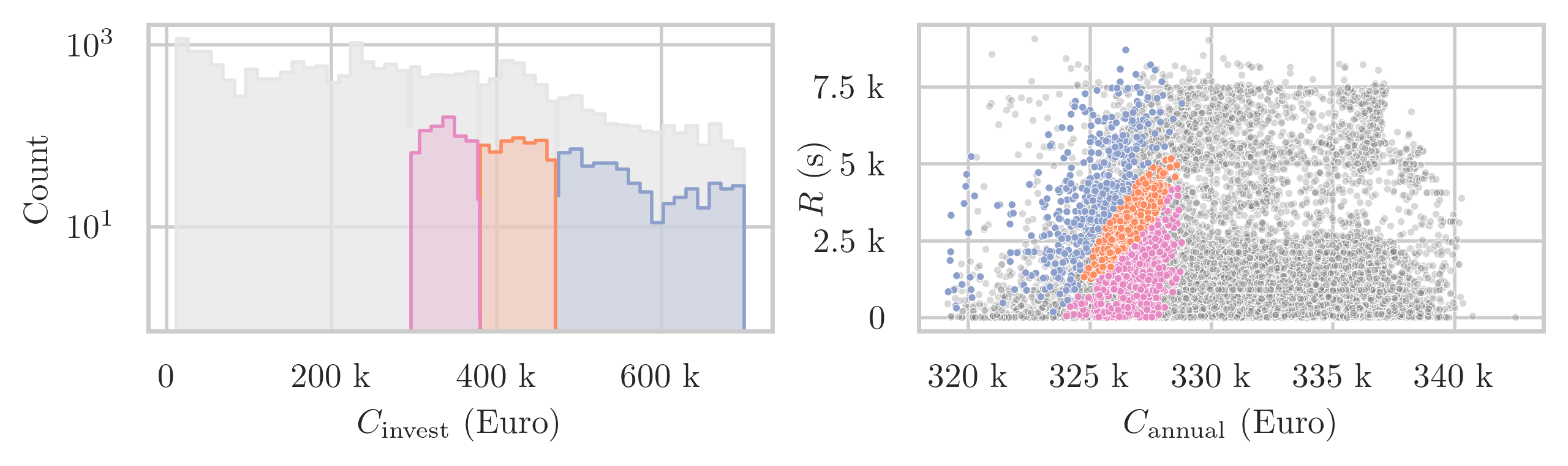}
\caption{
Secondary refinement step: The colored proportions and samples form the configuration concepts based on only the high investment cost samples from concept 3 from experiment 2.
} \label{fig:core_samples_DS_post_b}
\end{figure*}

The identified concepts are clearly separable with respect to the investment cost and demonstrate the achievable trade-off behavior for annual cost and resilience (Fig.~\ref{fig:core_samples_DS_post_b}).
Generally, it can be observed that higher investment allows for better resilience values given the similar or even lower annual cost.
Thus, this further processing step enables the desired fine-grained analysis of the high investment cost configurations and allows for a well-informed decision.

\section{Discussion}\label{sec:discussion}
The experiments demonstrate that concept identification produces meaningful and reasonable groups of energy management configuration options that are technically feasible and economically valid.
Technically, the approach maximizes a concept quality metric using a numerical optimization procedure. 
Due to the complexity of the optimization problem, the result is sensitive to small variations in the setup, such as initial conditions or choice of the optimizer~\cite{Lanfermann2020}, and thus most likely only represents local minima of the optimization problem.
In addition, the quality metric operates on technical aspects of the concept distribution, like size and overlap of the concepts, and incorporates the usefulness or desired trade-off relations only indirectly via the user-defined preference samples. 
As a result of both these aspects, the concepts identified in a specific setup are not unique and also each concept as a whole does not necessarily need to make sense to the decision maker. 
The latter is the case for concept 2 (green) in experiment 2, where, apart from representing low investment cost and large annual cost, multiple different types of configurations were sub-summarized in this concept. 
Therefore, the best way to utilize the concept identification method is in an iterative workflow where multiple concept identification processes are chained together, illuminating different aspects of the decision making problem in each analysis step.

The experiments of this work show how the choice of description spaces impacts the energy management configuration concepts derived from the identification process.
While the influence of the partitioning of the feature set into description spaces is generally not straightforward and non-intuitive to some extent, there are some insights on the effect available: 
Putting features into the same description space allows for each combination of feature values to be represented as a separate concept, regardless of other concepts (given that the concepts do not directly overlap). 
On the other hand, putting features into separate description spaces results in a tendency for feature values to have correlations, as only certain subsets of feature combinations are possible to be realized simultaneously. 
However, which concepts will be finally identified depends on many factors such as the number of desired concepts, the number and dimension of the several description spaces, the user preferences specified by user-defined samples, the details of the optimization process, and of course also the structure of the data set itself.

In experiment 1, a simple division into investment cost (description space 1) and the combination of annual cost and resilience (description space 2) leads to concepts that are separable (and ranked) based on investment cost and---at the same time---provide trade-off options for the other two objectives.
The first experiment produces valid configuration concepts, though neglects certain issues of technical feasibility and economic sensibility.
These aspects are integrated into the setup of experiment 2.
The allocation of two objectives into the same description space steers the identification process towards trade-offs for them.
A separation into different description spaces introduces a correlation requirement for these particular feature values.
This way, the process can, for example, identify a concept of low investment cost that effectively utilizes the stationary battery for peak power shaving (purple concept 1).
The third experiment illustrates a refinement process for the concept of high investment cost (yellow). 
This concept covers a wide range of configurations and for a well-motivated decision in this region a more detailed analysis is desirable. 
For the subsequent analysis, the resilience is determined to be an informative objective and therefore, the same setup of the description spaces as in experiment 1 is chosen. 
Based on the separation into investment cost (description space 1), as well as annual cost and resilience (description space 2), the high investment cost concept from experiment 2 is divided into three sub-concepts that are themselves ranked with respect to the investment cost and provide trade-off options for annual cost and resilience. 
As a result, the decision maker is now in a good position to arrive at a decision for a configuration. 
Of course, visualizations of the finally identified concepts in other description spaces or post-processing a selected subset of data again is entirely possible and for complex decisions surely warranted. 
But for the demonstrative purpose of this work, we refrain from showing more such in-depth analysis here.

\section{Conclusions}\label{sec:conclusion}

The present work studies the task of selecting viable energy management configurations from a complex data set consisting of \SI{20000}{} Pareto-optimal solutions from a ten-objective building energy management configuration problem.
We employ the concept identification approach to uncover semantically meaningful groups of solutions (concepts) which highlight trade-offs and design options in the decision making process. 
The proposed iterative work-flow of multiple concatenated concept identification processes allows the decision maker to generate refined insights and to introduce individual expectations and preferences.
We particularly focus on the influence of the choice of the description spaces, which can illuminate several different aspects of the data.
The series of experiments shows that the proposed approach can provide valuable insights into the engineering task and at the same time into the economic reasoning behind the configuration problem of energy management systems.

Future work could focus on defining a procedure to automatically assign features of a data set to description spaces, thereby increasing the accessibility of the method, independent of the availability of domain knowledge.
Technically, it would be desirable to formulate a constructive approach for the concept identification procedure which can generate concepts with high quality directly, without relying on cost-intensive black-box optimization.

The current study focuses on the general capabilities of the concept identification technique in the context of energy management systems. 
A necessary next step is to evaluate the quantitative improvements of the solutions found in the proposed procedure for a concrete application use-case. 
For example, how much do the found solutions meet the decision makers expectations for a concrete budget and set of design constraints?

We believe that the proposed methods can provide valuable insights in other application problems within the energy management domain. 
For example, the task of designing a mobility service comprising a ride-sharing capability with vehicle charging stations, charging schedule optimization, and battery management systems is very complex and could massively benefit from the application of concept identification.


\section*{Nomenclature}
\begin{description}[leftmargin=!,labelwidth=\widthof{\(C_\text{annual}\) }] 
  \item [$\alpha_{PV}$] Inclination angle of PV system
  \item [{\(\bar{b}\)}] Mean battery state of charge
  \item [{\(\beta_{PV}\)}] Orientation angle of PV system
  \item [{\(b_\mathrm{max}\)}] Maximum Battery state of charge
  \item [{\(b_\mathrm{min}\)}] Minimum Battery state of charge
  \item [{\(C_\text{annual}\)}] Annual operation cost
  \item [{\(C_\text{invest}\)}] Investment cost
  \item [{\(C_b\)}] Nominal Battery Capacity
  \item [{\(E_{d}\)}] Yearly discharged energy
  \item [{\(E_{f}\)}] Yearly feed-in energy
  \item [{\(G\)}] Yearly CO\textsubscript{2} emissions
  \item [{\(P_{c}\)}] Battery Charging Threshold
  \item [{\(P_{d}\)}] Battery Discharging Threshold
  \item [{\(P_{f}\)}] Maximum feed-in power peak
  \item [{\(P_{PV}\)}] Peak power of PV system
  \item [{\(P_{p}\)}] Maximum power peak
  \item [{\(Q\)}] Concept quality metric
  \item [{\(R\)}] Resilience
  \item [{\(t\)}] Medium SOC time share
  \item [{\(V\)}] Heat storage cylinder volume
\end{description}


\section*{Glossary}
\begin{description}[leftmargin=!,labelwidth=\widthof{MCDMM}] 
\item[CHP] Combined heat and power plant
\item[MDCM] Multi-criteria decision making
\item[PV] Photovoltaic
\item[SOC] State of Charge
\end{description}

\section*{Acknowledgments}
This work was supported in part by the National Natural Science Foundation of China under Grant No. 62302147 and No.62136003.



\begin{thebibliography}{10}
\expandafter\ifx\csname url\endcsname\relax
  \def\url#1{\texttt{#1}}\fi
\expandafter\ifx\csname urlprefix\endcsname\relax\def\urlprefix{URL }\fi
\expandafter\ifx\csname href\endcsname\relax
  \def\href#1#2{#2} \def\path#1{#1}\fi

\bibitem{SHAO2020377}
M.~Shao, Z.~Han, J.~Sun, C.~Xiao, S.~Zhang, Y.~Zhao, A review of multi-criteria
  decision making applications for renewable energy site selection, Renewable
  Energy 157 (2020) 377--403.

\bibitem{iijima2022automated}
F.~Iijima, S.~Ikeda, T.~Nagai, Automated computational design method for energy
  systems in buildings using capacity and operation optimization, Applied
  Energy 306 (2022) 117973.

\bibitem{rodemann2018many}
T.~Rodemann, A many-objective configuration optimization for building energy
  management, in: 2018 IEEE Congress on Evolutionary Computation (CEC), 2018,
  pp. 1--8.

\bibitem{faccio2018state}
M.~Faccio, M.~Gamberi, M.~Bortolini, M.~Nedaei, State-of-art review of the
  optimization methods to design the configuration of hybrid renewable energy
  systems (hress), Frontiers in Energy 12 (2018) 591--622.

\bibitem{gruber2015advanced}
J.~K. Gruber, F.~Huerta, P.~Matatagui, M.~Prodanovi{\'c}, Advanced building
  energy management based on a two-stage receding horizon optimization, Applied
  Energy 160 (2015) 194--205.

\bibitem{bui2019internal}
V.-H. Bui, A.~Hussain, Y.-H. Im, H.-M. Kim, An internal trading strategy for
  optimal energy management of combined cooling, heat and power in building
  microgrids, Applied Energy 239 (2019) 536--548.

\bibitem{nagpal2021optimal}
H.~Nagpal, I.-I. Avramidis, F.~Capitanescu, P.~Heiselberg, Optimal energy
  management in smart sustainable buildings--a chance-constrained model
  predictive control approach, Energy and Buildings 248 (2021) 111163.

\bibitem{delgarm2016multi}
N.~Delgarm, B.~Sajadi, F.~Kowsary, S.~Delgarm, Multi-objective optimization of
  the building energy performance: A simulation-based approach by means of
  particle swarm optimization (pso), Applied energy 170 (2016) 293--303.

\bibitem{harkouss2018multi}
F.~Harkouss, F.~Fardoun, P.~H. Biwole, Multi-objective optimization methodology
  for net zero energy buildings, Journal of Building Engineering 16 (2018)
  57--71.

\bibitem{he2019investment}
Y.~He, N.~Liao, J.~Bi, L.~Guo, Investment decision-making optimization of
  energy efficiency retrofit measures in multiple buildings under financing
  budgetary restraint, Journal of cleaner production 215 (2019) 1078--1094.

\bibitem{Liu2022}
Q.~Liu, F.~Lanfermann, T.~Rodemann, M.~Olhofer, Y.~Jin, Surrogate-assisted
  many-objective optimization of building energy management, IEEE Computational
  Intelligence Magazine (2023).

\bibitem{ciardiello2020multi}
A.~Ciardiello, F.~Rosso, J.~Dell'Olmo, V.~Ciancio, M.~Ferrero, F.~Salata,
  Multi-objective approach to the optimization of shape and envelope in
  building energy design, Applied energy 280 (2020) 115984.

\bibitem{triantaphyllou2000multi}
E.~Triantaphyllou, Multi-criteria decision making methods, Springer, 2000.

\bibitem{bejarano2023_clustering}
L.~A. Bejarano, H.~E. Espitia, C.~E. Montenegro, Clustering analysis for the
  pareto optimal front in multi-objective optimization, Computation 10~(3)
  (2022).

\bibitem{9209056}
Z.~He, G.~G. Yen, J.~Ding, Knee-based decision making and visualization in
  many-objective optimization, IEEE Transactions on Evolutionary Computation
  25~(2) (2021) 292--306.

\bibitem{Lanfermann2020}
F.~Lanfermann, S.~Schmitt, S.~Menzel, {An Effective Measure to Identify
  Meaningful Concepts in Engineering Design optimization}, in: 2020 IEEE
  Symposium Series on Computational Intelligence (SSCI), 2020, pp. 934--941.

\bibitem{Lanfermann2022a}
F.~Lanfermann, S.~Schmitt, Concept identification for complex engineering
  datasets, Advanced Engineering Informatics 53 (2022).

\bibitem{jing2019comparative}
R.~Jing, M.~Wang, Z.~Zhang, J.~Liu, H.~Liang, C.~Meng, N.~Shah, N.~Li, Y.~Zhao,
  Comparative study of posteriori decision-making methods when designing
  building integrated energy systems with multi-objectives, Energy and
  Buildings 194 (2019) 123--139.

\bibitem{schmitt2022incorporating}
T.~Schmitt, M.~Hoffmann, T.~Rodemann, J.~Adamy, Incorporating human preferences
  in decision making for dynamic multi-objective optimization in model
  predictive control, Inventions 7~(3) (2022) 46.

\bibitem{sedighizadeh2018stochastic}
M.~Sedighizadeh, M.~Esmaili, N.~Mohammadkhani, Stochastic multi-objective
  energy management in residential microgrids with combined cooling, heating,
  and power units considering battery energy storage systems and plug-in hybrid
  electric vehicles, Journal of Cleaner Production 195 (2018) 301--317.

\bibitem{yu2022extracting}
M.~G. Yu, G.~S. Pavlak, Extracting interpretable building control rules from
  multi-objective model predictive control data sets, Energy 240 (2022) 122691.

\bibitem{Lanfermann2022b}
F.~Lanfermann, S.~Schmitt, P.~Wollstadt, Understanding concept identification
  as consistent data clustering across multiple feature spaces, in: 2022 IEEE
  International Conference on Data Mining Workshops (ICDMW), 2022, pp.
  180--189.

\bibitem{Rosch1975}
E.~Rosch, {Cognitive reference points}, Cognitive Psychology 7~(4) (1975)
  532--547.

\bibitem{leImp2013}
M.~N. Le, Y.~S. Ong, S.~Menzel, Y.~Jin, B.~Sendhoff, {Evolution by Adapting
  Surrogates}, Evolutionary Computation 21~(2) (2013) 313--340.

\bibitem{Sim2013}
K.~Sim, V.~Gopalkrishnan, A.~Zimek, G.~Cong, {A survey on enhanced subspace
  clustering}, Data Mining and Knowledge Discovery 26~(2) (2013) 332--397.

\bibitem{Bickel2004}
S.~Bickel, T.~Scheffer, {Multi-View Clustering}, in: Fourth IEEE International
  Conference on Data Mining (ICDM'04), 2004, pp. 19--26.

\bibitem{dhillon2001_co-clustering}
I.~S. Dhillon, Co-clustering documents and words using bipartite spectral graph
  partitioning, in: Proceedings of the seventh ACM SIGKDD international
  conference on Knowledge discovery and data mining, 2001, pp. 269--274.

\bibitem{dhillon2003_information}
I.~S. Dhillon, S.~Mallela, D.~S. Modha, Information-theoretic co-clustering,
  in: Proceedings of the ninth ACM SIGKDD international conference on Knowledge
  discovery and data mining, 2003, pp. 89--98.

\bibitem{mirkin1996_clustering}
B.~Mirkin, {Mathematical Classification and Clustering}, Vol.~11 of Nonconvex
  Optimization and Its Applications, Springer US, Boston, MA, 1996.

\bibitem{Mechelen2004_two_mode_clustering}
I.~V. Mechelen, H.-H. Bock, P.~D. Boeck, Two-mode clustering methods: A
  structured overview, Statistical Methods in Medical Research 13~(5) (2004)
  363--394, pMID: 15516031.

\bibitem{Hartigan1972_direct_clustering}
J.~A. Hartigan, {Direct Clustering of a Data Matrix}, Journal of the American
  Statistical Association 67~(337) (1972) 123--129.

\bibitem{govaert2008_block_clustering}
G.~Govaert, M.~Nadif, Block clustering with bernoulli mixture models:
  Comparison of different approaches, Computational Statistics \& Data Analysis
  52~(6) (2008) 3233--3245.

\bibitem{schwanDigitalTwin2019}
T.~Schwan, S.~Schmitt, A.~Castellani, Calibration of hvac system models with
  monitoring data - digital twin meets measurement data, in: ESI FORUM IN
  DEUTSCHLAND, ESI, 2019.

\bibitem{rvmm}
Q.~Liu, R.~Cheng, Y.~Jin, M.~Heiderich, T.~Rodemann, Reference vector-assisted
  adaptive model management for surrogate-assisted many-objective optimization,
  IEEE Transactions on Systems, Man, and Cybernetics: Systems 52~(12) (2022)
  7760--7773.

\bibitem{chugh2016surrogate}
T.~Chugh, Y.~Jin, K.~Miettinen, J.~Hakanen, K.~Sindhya, A surrogate-assisted
  reference vector guided evolutionary algorithm for computationally expensive
  many-objective optimization, IEEE Transactions on Evolutionary Computation
  22~(1) (2016) 129--142.

\bibitem{Hao2022}
H.~Hao, A.~Zhou, H.~Qian, H.~Zhang, Expensive multiobjective optimization by
  relation learning and prediction, IEEE Transactions on Evolutionary
  Computation (2022).

\bibitem{Unger2012}
R.~Unger, B.~Mikoleit, T.~Schwan, B.~Bäker, C.~Kehrer, T.~Rodemann, Green
  building-modeling renewable building energy systems with emobility using
  modelica, in: Proceedings of Modelica 2012 Conference, Modelica Association,
  2012, pp. 897--906.

\bibitem{hansenCMA}
N.~Hansen, The cma evolution strategy: a comparing review, Towards a new
  evolutionary computation: Advances in the estimation of distribution
  algorithms (2006) 75--102.

\end{thebibliography}

\end{document}